%% file: sample_based_algo_team.tex
\documentclass[10pt,conference]{ieeeconf}

\usepackage{comment}
\usepackage[noend]{algpseudocode}
\usepackage{multicol}

\usepackage{authblk}
\usepackage{setspace}
\usepackage{bbm}
\usepackage{adjustbox}
\usepackage{longtable} 
\usepackage{xcolor} 
\usepackage{soul}
\usepackage{cite}
\usepackage{etoolbox}

\usepackage{kz_style}

\newcommand{\cib}{\pi}

\newcommand{\com}{c}
\newcommand{\vir}{h}
\newcommand{\gro}{\gamma_\text{rollout}}
\newcommand{\Gro}{\Gamma_\text{rollout}}

\newcommand{\loc}{m}
\newcommand{\locset}{\mathcal{M}}

\newcommand{\Npre}{\mathcal{N}^-}
\newcommand{\Npost}{\mathcal{N}^+}
\newcommand{\presc}{\gamma}
\newcommand{\Presc}{\Gamma}
\newcommand{\protloc}{\cP_L}
\newcommand{\protcom}{\cP_Z}
\newcommand{\uv}{u}
\newcommand{\uset}{\cU}

\newcommand{\xv}{x}
\newcommand{\wv}{w}
\newcommand{\xset}{\cX}

\newcommand{\yv}{y}
\newcommand{\yset}{\cY}

\newcommand{\kznote}[1]{\vspace{1em}\noindent{\leavevmode\color{blue}[Kaiqing -- #1]\newline}}

\newcommand{\emnoteinline}[1]{{\color{magenta}[Erik -- #1]}}
\newcommand{\note}[1]{\vspace{1em}\noindent{\leavevmode\color{cyan}[Note -- #1]\newline}}

\newcommand{\issue}[1]{\vspace{1em}\noindent{\leavevmode\color{red}[{\bf Issue} -- #1]\newline}}
\newcommand{\delete}[1]{\vspace{1em}\noindent{\leavevmode\color{red}[Delete -- #1]\newline}}
\newcommand{\edit}[1]{{\color{brown}#1}}

\title{\LARGE \bf
Online Planning for Decentralized Stochastic  Control \\ with Partial History Sharing}

\author{Kaiqing~Zhang, Erik~Miehling, and Tamer~Ba{\c s}ar\thanks{This work was  supported in part by the U.S. Office of Naval Research (ONR) MURI grant
N00014-16-1-2710, and in part by the US Army Research Office (ARO)
Grant W911NF-16-1-0485.  The authors  are with the Coordinated Science Laboratory, University of Illinois at Urbana-Champaign (\{kzhang66, miehling, basar1\}@illinois.edu). }}

\IEEEoverridecommandlockouts

\begin{document} 

\maketitle  

\begin{abstract}

In decentralized stochastic control, standard approaches for sequential decision-making, \emph{e.g.} dynamic programming, quickly become intractable due to the need to maintain a complex information state. Computational challenges are further compounded if agents do not possess complete model knowledge. In this paper, we take advantage of the fact that in many problems agents share some common information, or history, termed \emph{partial history sharing}. Under this information structure the policy search space is greatly reduced. We propose a provably convergent, online tree-search based algorithm that does not require a closed-form model or explicit communication among agents. Interestingly, our algorithm can be viewed as a generalization of several existing heuristic solvers for decentralized partially observable Markov decision processes. To demonstrate the applicability of the model, we propose a novel \emph{collaborative intrusion response} model, where multiple agents (defenders) possessing asymmetric information aim to collaboratively defend a computer network. Numerical results demonstrate the performance of our algorithm. 
\end{abstract}

\input{Introduction}

\input{Modeling}

\input{Algorithm}

\input{Analysis}

\input{Example}

\input{DiscussionAndConclusion}

\bibliographystyle{IEEEtran}  
\bibliography{RL,StochasticControl}

\end{document}

%% file: Introduction.tex
\section{Introduction}\label{sec:Intro}

{
The lack of centralized information in decentralized control settings introduces significant theoretical and computational challenges. The primary difficulty arises from the fact that no agent knows all past system information (consisting of past actions and observations, termed the \emph{history}) that is relevant to its decision making process, a property referred to as \emph{perfect recall} \cite{nayyar2011optimal}. 
In centralized control settings, perfect recall allows for the application of the \emph{one-way separation between estimation and control} \cite{witsenhausen1971separation} -- {the system state estimate, given past system information, is independent of the previous decision rules (the functions that map the available information into control actions), \emph{i.e.}, policy-independence.  }
This results in the reduction of the history to a \emph{sufficient statistic}\footnote{The information that is sufficient for making optimal decisions, also termed an \emph{information state} or \emph{belief}.}, the probability distribution over states, that is independent of the choice of past decision rules (see \cite{striebel1965sufficient,kumar1986stochastic}) and thus has a time-invariant domain. Importantly, the update of the sufficient statistic  is a function of the control action, and not the entire decision rule. 
In decentralized control settings, no agent has perfect recall and thus the separation principle cannot be applied. Asymmetric information among agents results in a more complex sufficient statistic and, due to the lack of perfect recall, requires one to take into account the influence of previous decision rules (instead of only the control action) on the update of the sufficient statistic and, in turn, on the computation of the optimal control.
}

The general problem of dynamic decision making when no agent possesses complete system knowledge has been studied in the literature spanning multiple fields. In the control theory literature, the problem is referred to as decentralized stochastic control, whereas in the computer science literature, the problem is {predominantly} studied under the topic of decentralized partially observable Markov decision processes (Dec-POMDPs). Generally speaking, algorithms developed to address the problem are subject to at least one of the following two  limitations. 
First, some algorithms are based on the assumption that the knowledge of the model is known either to all agents \cite{nayyar2013decentralized} or to a central coordinator that designs policies for all agents \cite{oliehoek2014dec}, which is usually not true in practice. 
Second, many of the algorithms require planning to be done \emph{offline}, \emph{i.e.}, \emph{bottom-up} approaches where the optimal policies of all agents are calculated before runtime via dynamic programming \cite{hansen2004dynamic,seuken07improvedmemory,oliehoek2013sufficient,dibangoye2016optimally}. Unfortunately, these exact approaches generate significant computational issues even for problems with a moderate dimension or planning horizon.  
In response, other work has resorted to approximations that rely on forward lookahead search (or \emph{top-down} approaches). Unfortunately, many of these algorithms are heuristic and are not guaranteed to converge to the optimal\footnote{{The term \emph{optimal} is taken to mean \emph{team-optimal} \cite{yuksel2013stochastic}, interpreted as the joint policy that maximizes the total reward given the informational constraints of the agents.} } policies \cite{oliehoek2014dec}. 


With the goal of addressing these limitations, we propose a provably convergent online planning algorithm for decentralized stochastic control which can be implemented in a decentralized fashion, without requiring any communication among agents.
Our  {proposed algorithm} exactly addresses two key challenges/drawbacks {of decentralized decision making} 
\cite{dibangoye2018learning}: the explosion in {complexity as a function of the state space}, and the centralization of computation. 
In particular, we first take advantage of the fact that in many problems decision makers share some {common} information, 
{what is referred to as \emph{partial history sharing} (PHS) in the literature  \cite{nayyar2013decentralized}}. For example, decision makers may observe each other's control actions (\emph{e.g.}  fleet control of self-driving cars  \cite{gerla2014internet}), or may share some common observations (\emph{e.g.} cooperative navigation of robots  \cite{zhang2018fully}). 
Following the structural results developed in \cite{nayyar2013decentralized},
 {utilizing common information} allows for the reformulation of the decentralized stochastic control problem {as} an {equivalent} centralized partially observable Markov decision process (POMDP). The state of the centralized POMDP includes both the underlying state of the dynamic system as well as the local information of the controllers. The space of information states of the centralized POMDP, \emph{i.e.}, the space of joint {probability distributions} over the system state and local information of agents, is greatly reduced compared to the \emph{multi-agent belief state} \cite{hansen2004dynamic,seuken07improvedmemory} {commonly used in the Dec-POMDP literature}.  Such a reduction further enables a {compression} of the policy search space (a nice argument of this \emph{compression} can be found in Section I-A of  \cite{nayyar2013decentralized}).
To solve the centralized problem  efficiently, we adopt a \emph{top-down} approach based on  Monte-Carlo tree search for solving large-scale POMDPs \cite{silver2010monte}.  
To enable a decentralized implementation, all agents construct a local copy of the search tree, and are assumed to share a common source of randomness, \emph{e.g.}, a common random number generator, so that the {agents'} generated trees are identical, {obviating the need for explicit communication}. 

We demonstrate the performance of the proposed algorithm in a computer security setting, focusing on how multiple agents (defenders) can collaboratively defend a cyber network subject to the constraint that they cannot instantaneously share defense actions and intrusion information {with others}. As   will be shown, the setting is naturally addressed by a form of PHS, termed the \emph{delayed sharing information structure}, allowing for direct application of the proposed algorithm.


\subsection{Related Work}\label{subsec:lit_review}

 

Most existing solvers for decentralized decision-making rely on a reformulation of the decentralized problem as a centralized problem \cite{nayyar2013decentralized,oliehoek2014dec,hansen2004dynamic,dibangoye2016optimally}. 
One prominent approach in the literature is the work of \cite{nayyar2013decentralized} in which the authors introduce a common information approach, and an associated dynamic programming decomposition, where agents are assumed to share some of their history with others. 
 Interestingly, the reformulation of \cite{nayyar2013decentralized} includes several approaches for Dec-POMDPs from the computer science community as special cases, when the common information is absent. In particular, the common information belief state in \cite{nayyar2013decentralized} generalizes the definition of \emph{occupancy-state MDP} in \cite{dibangoye2016optimally,dibangoye2018learning}, which was shown to be the sufficient statistic {for}  solving general  Dec-POMDPs.

To solve the centralized problem efficiently, several \emph{top-down}  algorithms have been proposed, \emph{e.g.}, \cite{oliehoek2014dec} converted the Dec-POMDP to a type of {centralized sequential decision problem, termed a \emph{non-observable MDP} (NOMDP)}, which was then solved by the heuristic \emph{MAA$^*$} tree search algorithm \cite{szer2005maa}.  
Several sampling-based planning/learning algorithms have also been proposed to improve the tractability of Dec-POMDP solvers. {In particular, }
\cite{wu2013monte} developed solvers that {combined}   Monte-Carlo sampling {with} policy iteration and {the expectation-maximization} algorithm, respectively. 
In addition, Monte-Carlo tree search   has been applied {to} special classes of Dec-POMDPs, \emph{e.g.},   multi-agent POMDPs \cite{amato2015scalable} and multi-robot active perception \cite{best2018dec}. {Under PHS, the only model-free learning algorithm that we are aware of is the work of \cite{arabneydi2015reinforcement}.}

Most existing planning/learning procedures  {are designed} to be implemented in a centralized fashion, \emph{i.e.}, either the coordinator  designs optimal policies for all agents \cite{oliehoek2014dec,hansen2004dynamic,dibangoye2018learning}, or each agent communicates with   other agents to access   global  information \cite{zhang2018fully,best2018dec,oliehoek2012tree,zhang2018networked,zhang2018finite}.  To enable a \emph{fully decentralized} implementation, we {assume that all agents} use a common random number generator  {when sampling distributions}.   {This idea, termed a \emph{correlation device}, has been used in the past to avoid explicit communication {between agents} \cite{arabneydi2015reinforcement,bernstein2009policy}}. Such correlation devices are reminiscent of similar devices used in generating correlated equilibria in games \cite{basar1999dynamic}. 



%

\subsection{Contribution}

{The contribution of our paper is} three-fold: {\bf 1)} We develop a tractable online planning algorithm for  decentralized stochastic control problems with partial history sharing. The algorithm does not need explicit  knowledge\footnote{Instead, the existence of a simulator/generative model is assumed.} of the model, requires no  explicit communication\footnote{As discussed in \cite{amato2013decentralized}, \emph{no communication} refers to no \emph{explicit} information transmission among agents. However, it is still possible for agents to share information via {modification and observation of} the underlying system.} between agents, and is provably convergent to the team-optimal solution. {\bf 2)} We offer a unifying perspective 
of existing algorithms through the lens of decentralized stochastic control. Specifically, we describe (in Section \ref{sec:connection}) how some recently proposed Dec-POMDP solvers   can be viewed as special cases of our algorithm. {\bf 3)} The proposed algorithm is applied to a novel computer security setting, which we term \emph{collaborative intrusion response.} To the best of our knowledge, this is the first decentralized approach to intrusion response.

We note that although such a common-information based  framework has been advocated in the control theory literature \cite{nayyar2013decentralized}, it has not been fully investigated in the context of developing  planning/learning algorithms for Dec-POMDPs. In this sense, our work offers a new perspective for improving the computational efficiency of solving Dec-POMDPs, specifically for problems where common information exists.


%

%% file: Modeling.tex
\section{Preliminaries}\label{sec:Modeling}
In this section, we introduce the model for decentralized stochastic control under partial history sharing, and review the {relevant structural results  from \cite{nayyar2013decentralized}.

\subsection{Decentralized Stochastic Control with PHS}\label{sec:Info_structure}
Consider a system of $n$ cooperative decision makers (hereafter referred to as agents) operating over a horizon of length $T$. At each time $t$, the system takes on one of finitely many states from the space $\xset$. The system state evolves as a function of the control actions of the agents: given a current {system} state and the collection of control actions across agents, {termed a \emph{joint control action}, denoted by $\uv=(\uv^1,\ldots,\uv^n)\in\uset=\uset^1\times\cdots\times\uset^n$, the {system} state obeys the following dynamics $\xv_{t+1} =f(\xv_t,\uv_t,\wv_t^0)$, where {$\{\wv_t^0\}_{t=1}^{T}$} is a sequence of independent and identically distributed (i.i.d.) random variables. 
Agents incur a common reward (since they are cooperative), denoted by $r(\xv,\uv)$, which is a function of the underlying {system} state $\xv$ and the joint control action $\uv$.}








{Each agent lacks perfect information of the underlying system state, instead receiving local (noisy) observations as the system evolves. This leads to agents possessing asymmetric information}. 
Here, we focus on a special while prevalent information structure, termed partial history sharing, where  {agents'} local information may have some components in common. 
In particular, at time $t$, three pieces of information are available to each agent: a local  observation, local memory, and the common history. 

\begin{itemize}
	\item \textbf{{Local}  observations}: Each agent $i$ receives {a local observation $\yv^i_t\in\cY^i$}, which is generated {according to} $\yv^i_t=h_t^i(\xv_t,\wv^i_t)$, where {$\{\wv^i_t\}_{t=1}^{T}$} is a sequence of i.i.d. random variables. {The collection of observations across all agents is denoted by $\yv_t=(\yv^1_t,\cdots,\yv^n_t)\in\yset = \yset^1\times\cdots\times\yset^n$}.
	\item \textbf{Local memory/information}: Each agent $i$ maintains a {local} memory $\loc^i_t\in \locset_t^i$, {a subset of $\{\uv^i_{1:t-1},\yv^i_{1:t}\}$, representing their} (possibly limited) storage of the local observations and actions {up to and including} time $t$. 
	\item \textbf{Common history/information}: In addition to the local memory, all agents possess 
a common history $\com_t\subseteq\{\uv_{1:t-1},\yv_{1:t}\}$, which {encodes any shared history} of the local observations and actions of all agents. 
\end{itemize}

Each agent $i$ makes decisions based on its currently available information {$(\loc^i_t,\com_t)$}. Formally, let $g^i_t$ denote the \emph{control law of agent $i$} that maps agent $i$'s current information $(\loc^i_t,\com_t)$ into a local control action $\uv^i_t$, that is
\#\label{equ:def_git}
\uv^i_t=g^i_t(\loc^i_t,\com_t).
\#
The collection of agent $i$'s control laws over time is called the \emph{control policy of agent $i$} {and is denoted by} ${g}^i=(g^i_1,\cdots,g^i_{T})$. {The collection of control policies across all agents is called the \emph{control policy of the system} and is denoted by ${g}=({g}^1,\cdots,{g}^n)$.}


{
The order of events in the model is now discussed.} For a given time-step $t$, each agent $i$ takes an action, $\uv_t^i$, receives a local observation, $\yv_{t+1}^i$, and then shares a subset $z^i_{t+1}$ of their (updated) local information $\{\loc_t^i,\uv_t^i,\yv_{t+1}^i\}$ with all other agents. 
The specific information that is shared is dictated by the function $\protcom^i$, that is $z^i_{t+1} = \protcom^i(m_t^i,u_t^i,y_{t+1}^i)\in\cZ^i$. {We term the quantity $z_{t+1}^i$ the \emph{innovation}, and define the joint innovation as $z_{t+1}=(z^1_{t+1},\cdots,z^n_{t+1})\in\cZ = \cZ^1\times\cdots\times\cZ^n$.} 
Each agent then updates their local memory according to the function $\protloc^i$, dictating what information {is carried over to} the next iteration, that is $\loc_{t+1}^i = \protloc^i(\loc^i_{t},\uv^i_t,\yv^i_{t+1},z^i_{t+1})\in\locset_{t+1}^i$.\footnote{The dependency of $\loc_{t+1}^i$ on $z^i_{t+1}$ is to ensure that {agent's} local information and the common information are disjoint, $\loc_{t+1}^i\cap c_{t+1}=\varnothing$.} Finally, the common history is updated as $\com_{t+1} = \{\com_t,z_{t+1}\}$. The functions {$\protcom^i$ and $\protloc^i$} are dictated by the problem setting at hand (see Section \ref{sec:Example} for their definition in the context of our computer security example).



The objective of the problem is to determine the control policy {$g$ 
that maximizes} the expected total discounted reward, defined as
\begin{align}\label{eq:totalreturn}
R(g) := {\EE^g\left[\sum_{t=1}^{T}\beta^tr(x_t,\uv_t)\right]}
\end{align}
where $r(\xv,\uv)$ is the instantaneous reward received when joint action $\uv$ is taken in system state {$x$}, $\beta\in[0,1]$ is the discount factor, 
and $\EE^g$ denotes {the expectation with respect to the probability measure induced by policy $g$}. 

\subsection{Common Information Based Approach}
\label{ssec:com}



{
In general, the decentralized problem with partial history sharing is equivalent to a centralized problem from the perspective of some \emph{virtual coordinator} {\cite{nayyar2013decentralized}}. The coordinator is assumed to  have access only to the information that is in common among all agents, that is, the common history described in Section \ref{sec:Info_structure}. The coordinator solves for functions for each agent, termed \emph{prescriptions}\footnote{Note that in the control and computer science communities, both terms, prescriptions and policies, have been used. We will use them interchangeably.}, that map  local information ({\emph{i.e.} local memories) to control actions.} 
It was shown in \cite{nayyar2013decentralized} that the coordinator's problem of determining these prescriptions reduces to a POMDP with appropriately defined state, action, and observation spaces. 


In particular, consider a coordinator that can observe the {common history} $\com_t$ but not the local observations $\yv^i_{t}$ or the local memories  $\loc^i_t$. {The coordinator's decision is the set of prescriptions across all agents, denoted by $\presc_t=(\presc^1_t,\cdots,\presc^n_t)$, where $\presc^i_t:\cM^i_t\to\cU^i$ is a mapping from the local memory to a local control action, that is, $\uv^i_t=\presc^i_t(\loc^i_t)$. The set of all prescriptions $\presc^i_t$ is denoted by $\Presc^i_t$ with $\Presc_t = \Presc_t^1\times\cdots\times\Presc_t^n$.} 
By Lemma $1$ of \cite{nayyar2013decentralized}, the {coordinator's} problem can be viewed as a POMDP with state $(x_t,\loc^1_t,\cdots,\loc^n_t)\in\xset\times\locset_t^1\times\cdots\times\locset_t^n$, {actions $\presc_t=(\presc^1_t,\cdots,\presc^n_t)\in\Presc_t$, and observations $z_t\in\cZ$}. 
By defining the \emph{virtual history} of the coordinator\footnote{To avoid confusion with the actual history of actions and observations in the original decentralized stochastic control problem, we use the term \emph{virtual history} to represent the history of actions ($\presc_t$) and observations ($z_t$) of the equivalent centralized POMDP (the virtual coordinator's problem).} as a sequence of joint prescriptions and innovations, that is $h_t=\{\presc_1,z_2,\presc_2,\cdots,\presc_{t-1},z_t\}$, one can define the information state, or belief state, for the centralized POMDP as the distribution over $(x_t,\loc^1_t,\cdots,\loc^n_t)$ given virtual history $h_t$, that is
\#\label{equ:def_info_state}
\pi_t=\PP(x_t,\loc^1_t,\cdots,\loc^n_t\mid h_t).
\#
%
The update of {the} information state $\pi_t$ follows the transition $\pi_{t+1}=\phi(\pi_t,\presc_t,z_t)$ for some function $\phi$ (cf. Appendix A in \cite{nayyar2013decentralized}).   
Moreover, the reward of the centralized POMDP is
\$
\tilde{r}(\pi_t,\presc_t)=\EE[r(x_t,u_t)\mid c_t,\presc_{1:t}]=\EE[r(x_t,u_t)\mid h_t,\presc_{t}],
\$
where $u^i_t=\presc^i_t(m^i_t)$. Note that the expectation is taken over the probability distribution $\pi_t$. The value function of the POMDP is defined as
\#\label{equ:def_value}
V^*_t(h_t)=\sup_{\mu_t,\cdots,\mu_T}\EE^{\mu_t,\cdots,\mu_T}\Bigg[\sum_{\tau=t}^T\beta^{\tau-t}\tilde{r}(\pi_\tau,\presc_\tau)\Bigg]
\#
where $\mu_t$, termed the \emph{coordination strategy}, is a mapping from the virtual history to prescriptions. The coordination strategy is analogous to the notion of a policy in conventional POMDPs. 
Note that $V^*_1(h_1)$ is exactly the maximum of the expected total discounted  reward defined in \eqref{eq:totalreturn}.

Given any optimal coordination strategy $\mu_t^*$, the optimal joint prescription is determined by $\presc_t^*=\mu_t^*(h_t)$ {where $\presc_t^*=(\presc_t^{1,*},\cdots,\presc_t^{n,*})$}. {Consequently}, by Theorem $2$ of \cite{nayyar2013decentralized}, 
the optimal control action for any agent $i$  is determined by } 
\begin{align}
\uv^{i,*}_t=
{\presc^{i,*}_t}(\loc^i_t). \label{eq:prescription}
\end{align} 
For finite action and memory spaces, the space of possible prescriptions is also finite, that is $|\Presc_t| = \prod_{i=1}^n{|\cU^i|}^{|\cM^i_t|}$, 
and thus the coordinator's functional optimization (of determining the optimal prescriptions) reduces to an optimization over vector actions. 






The POMDP reformulation enables {a sequential decomposition of the problem and thus the construction of a backward induction algorithm, via dynamic programming,} for finding an optimal control policy \cite{nayyar2013decentralized}. However, the backward induction algorithm requires solving a {sequence of} one-stage functional optimization problems for all realizations of $\pi_t$, {which is computationally challenging given that $\pi_t$ lives in an infinite dimensional space.} 
The aim of our algorithm (introduced next) is to avoid this computational burden by selecting prescriptions for the current information state $\pi_t$ via online construction of search trees.



%% file: Algorithm.tex
\section{Decentralized Online Planning}\label{sec:Alg}

In this section, we outline our online algorithm for solving the decentralized stochastic  control problem {with} partial history sharing. The algorithm is inspired by the single-agent (centralized) POMDP algorithm known as \emph{partially observable Monte-Carlo planning} (POMCP) \cite{silver2010monte}. Using the reduction of the decentralized control problem to a centralized POMDP shown in Section \ref{ssec:com}, the development of a decentralized algorithm based on a single-agent algorithm is fitting. 
Since the coordinator's problem is based on common information, all agents know this information and can individually solve the coordinator's problem, thus obtaining a solution to the original decentralized control problem. 

\subsection{The Search Tree of Virtual Histories}


As in POMCP \cite{silver2010monte}, the basis of our algorithm is a search tree constructed iteratively via Monte-Carlo simulations (described in more detail in Section \ref{ssec:dec}). One key difference here is that nodes in the search tree correspond to the virtual histories $h_t$, as defined in Section \ref{ssec:com}, instead of the histories of actions and observations as in the POMCP algorithm. The search tree, denoted by $\cT$, consists of nodes, denoted by $\cT(h)$, each of which encodes two quantities $\cT(h) = (N(h),V(h))$: a count index $N(h)$, describing how many times node $h$ has been visited in past simulations, and an estimated value $V(h)$, representing the mean value of all simulations that began at virtual history $h$.\footnote{For brevity, we do not include a subscript $t$ for the estimated value $V(h)$. The time index $t$ will be self-evident from the argument $h$.} 
For a given node $h$, each branch emanating from $\vir$ is an alternating sequence of joint prescriptions $\delta=(\delta^1,\ldots,\delta^n)\in\Presc$ and joint innovations $z\in\cZ$. 

\subsection{Algorithm: Decentralized {Online} Planning with PHS}
\label{ssec:dec}


We now describe our proposed algorithm, termed \emph{decentralized online planning with partial history sharing}. A fundamental characteristic of our algorithm is that computation is decentralized, that is, we do not rely on a centralized entity to compute agents' control policies. To this end, each agent $i$ constructs its own copy of the search tree, denoted by $\cT^i$. To carry out simulations, we assume that each agent has access to a \emph{generative model} (a black-box simulator), $\cG$, that takes as input a system state $x$ and joint action $(u^1,\ldots, u^n)$, and outputs a successor system state $x'$, joint observation vector $(y^1,\ldots,y^n)$, and reward $r$. The generate model avoids the need for an explicit model representation. 


The algorithm consists of two main stages: the search stage and a belief update stage. In the search stage, each agent begins their simulations from the same virtual history $\vir$. 
%
%
%
Each agent draws a sample $(\xv,\loc^1,\ldots,\loc^n)$ from the current belief, approximated by a set of particles $B(\vir)$. 
Using this sample, each agent expands the search tree from the root node $\vir$ using either a \emph{rollout} simulation, in the case where $\vir$ does not have children nodes, or a selection rule (UCB1 \cite{auer2002finite}) if $\vir$ already has children. The UCB1 selection rule balances exploration and exploitation by maximizing the sum of the current estimated value of prescription $\delta$, $V(\vir\delta)$, and an exploration term $\rho\sqrt{\frac{\log N(\vir)}{N(\vir\delta)}}$ depending on the number of times $h$ has been visited, $N(\vir)$, and the number of times $\delta$ has been selected from $h$, $N(\vir\delta)$. Successive simulations further expand the search tree and, due to the above selection rule, allow for targeted search of the decision space and efficient convergence of estimates. The pseudocode of the search stage, for a given agent $i$, is shown in Algorithm \ref{Alg1}.

To enable a decentralized algorithm, we assume that agents' samples are correlated via a common source of randomness. Practically, this means that any time in the algorithm an agent draws a sample from a distribution, it is done so using a pseudorandom number generator with a common random seed. As a result, agents construct identical search trees and compute the same optimal joint prescription. Since agents are completely cooperative, they can agree upon this common seed beforehand and, further, each agent $i$ can rely on every other agent $j$ to follow the prescription function that was computed {during} the search stage.


Under the computed joint prescription, each agent $i$ uses its local information to specify an action, and transmits {their} innovation $z_{t+1}^i$ to all other agents using the function $\protcom^i$. Local memories are then updated via the function $\protloc^i$, relevant branches of the search tree are identified, and the virtual history is updated allowing for the next round of the search algorithm to proceed.

Upon update of the virtual history, each agent must update its belief $\pi$ in order to reflect the new information ($\presc_t,z_{t+1}$). In practice, it is not tractable to maintain an exact belief representation. Instead, for a given virtual history $\vir_t$, each agent's belief $B(h_t)$ is represented as a set of $K$ particles of the form $B^j_t\in\xset\times\locset_t^1\times\cdots\times\locset_t^n$.\footnote{The specific approximation is $\frac{1}{K} \sum_{j=1}^K\delta_{(x,\loc^1,\ldots,\loc^n),B^j_t}$, where $\delta_{\cdot,\cdot}$ is the Kronecker delta function.} Each agent updates its belief by drawing a sample $(\xv,\loc^1,\ldots,\loc^n)$ from the current belief approximation $B(\vir)$ then, using the computed prescription, specifies a joint action $(\uv^1,\ldots,\uv^n)$. The generative model is then called to obtain a sample $(x',y^1,\ldots,y^n,-)$ which is used to construct a {joint} innovation and an updated set of local memories $(\loc^1{'},\ldots,\loc^n{'})$. If the sampled {joint} innovation matches the true  {joint} innovation, then the particle $(\xv',\loc^1{'},\ldots,\loc^n{'})$ is added to $B(h')$. The sampling repeats until $K$ particles have been accepted into $B(h')$. {The common source of randomness results in agents possessing identical updated belief approximations}.

\begin{algorithm*}

  \caption{Decentralized Online Planning with Partial History Sharing -- Agent $i$}\label{Alg1}
  \vspace{0em}
  \begin{spacing}{1.15}
  \small
    \hspace{-0.025\textwidth}
    \begin{minipage}[t]{0.49\textwidth}
      \begin{algorithmic}[0]
        \Function{Search}{$\vir$}
	\Repeat
	\If{$\vir=\varnothing$}: $(x,\loc^1,\ldots,\loc^n)\sim B_0$
	\Else: $(x,\loc^1,\ldots,\loc^n)\sim B(\vir)$
	\EndIf
	\State $\textsc{Simulate}(x,\loc^1,\ldots,\loc^n,\vir,0)$
	\Until \textsc{StoppingCondition}()
	\State\Return $\argmax_{\delta\in\Gamma} V(\vir\delta)$
        \EndFunction  
      \end{algorithmic}
      \vspace{0em}
      \begin{algorithmic}[0]
        \Function{Rollout}{$x,\loc^1,\ldots,\loc^n,\vir,{d}$}
        \If{$\beta^d<\varepsilon$}: \Return $0$
        	\EndIf
	\State $\displaystyle\presc=(\presc^1,\ldots,\presc^n)\sim(\Gro^1(\vir),\ldots,\Gro^n(\vir))$
        \State $(u^1,\ldots,u^n)\leftarrow(\presc^1(\loc^1),\ldots,\presc^n(\loc^n))$
        \State $(x',y^1,\ldots,y^n,r)\sim\cG(x,u^1,\ldots,u^n)$
        \State $z^i\leftarrow \protcom^i(\loc^i,u^i,y^i)$ and share $z^i$ with all other agents
        \State $\vir'\leftarrow \vir\presc z$
        \State $(\loc^1{'},\ldots,\loc^n{'})\leftarrow(\protloc^1(\loc^1,u^1,y^1,z^1),\ldots,$
        \State $\hspace{13em}\protloc^n(\loc^n,u^n,y^n,z^n))$
	\State $R\leftarrow r+\beta\cdot\textsc{Rollout}(x',\loc^1{'},\ldots,\loc^n{'},\vir',d+1)$
        \State\Return $R$
        \EndFunction  
      \end{algorithmic}
    \end{minipage}
    \hspace{-0.025\textwidth}
    %
    \begin{minipage}[t]{0.49\textwidth}
    \begin{algorithmic}[0]
        \Function{Simulate}{$x,\loc^1,\ldots,\loc^n,\vir,d$}
        \If{$\beta^d<\varepsilon$}: \Return $0$
        	\EndIf
        \If{$\vir\not\in\cT^i$}
        \ForAll{$\presc\in\Presc$}
        \State $\cT^i(\vir\presc)\leftarrow (N_0(\vir\presc),V_0(\vir\presc))$
        \EndFor
        \State\Return \textsc{Rollout}($x,\loc^1,\ldots,\loc^n,\vir,d$)
        \EndIf
        \State $\presc\hspace{-0.25em}=\hspace{-0.15em}\displaystyle(\presc^{1},\ldots,\presc^{n})\in\hspace{-0.25em}\argmax_{(\delta^1,\ldots,\delta^n)\in\Presc^1\times\cdots\times\Presc^n}\hspace{-1.25em}V(\vir\delta)+\rho\sqrt{\frac{\log N(\vir)}{N(\vir\delta)}}$ \vspace{0.425em}
        \State $(u^1,\ldots,u^n)\leftarrow(\presc^{1}(\loc^1),\ldots,\presc^{n}(\loc^n))$
        \State $(x',y^1,\ldots,y^n,r)\sim\cG(x,u^1,\ldots,u^n)$
        \State $z^i\leftarrow \protcom^i(\loc^i,u^i,y^i)$ and share $z^i$ with all other agents
        \State $\vir'\leftarrow \vir\presc z$
        \State $(\loc^1{'},\ldots,\loc^n{'})\leftarrow(\protloc^1(\loc^1,u^1,y^1,z^1),\ldots,\protloc^n(\loc^n,u^n,y^n,z^n))$
        \State $N(\vir)\leftarrow N(\vir)+1$
        \State $R\leftarrow r+\beta\cdot\textsc{Simulate}(x',\loc^1{'},\ldots,\loc^n{'},\vir',d+1)$ 
        \State $N(\vir\presc)\leftarrow N(\vir\presc)+1$
        \State $V(\vir\presc)\leftarrow V(\vir\presc) + \frac{R-V(\vir\presc)}{N(\vir\presc)}$
        \State\Return $R$
        \EndFunction
      \end{algorithmic}
    \end{minipage} 
    \end{spacing}
    \vspace{1em}
\end{algorithm*}

%% file: Analysis.tex
\subsection{Convergence}\label{sec:Analys}


{By virtue of the common source of randomness, the planning update, \emph{i.e.}, the construction of the search trees, is identical and decoupled for all agents.  As a result, the convergence of our decentralized online planning algorithm can be characterized by that of the single-agent POMCP algorithm \cite{silver2010monte}. The convergence of our algorithm is given as follows. }

{
\begin{lemma}[{Theorem} $2$ \cite{silver2010monte}]\label{lemma:conv_prop}
	{Given the true belief state $\pi_\tau$}, the value function constructed by Algorithm \ref{Alg1} converges
 in probability to the  value function, \emph{i.e.}, $V(h_\tau)\xrightarrow{p} V^*_\tau(h_\tau)$, for any history  $h_\tau$ that are prefixed by $h_t$ with $\tau\geq t$, where $V^*_\tau(h_\tau)$ is as defined in \eqref{equ:def_value}. As the number of visits $N(h_\tau)$ approaches infinity, the bias of the value function, $\EE[V(h_\tau)-V^*_\tau(h_\tau)]$, reduces on{ the order of $\log N(h_\tau)/N(h_\tau)$.} 
\end{lemma}
}

Lemma \ref{lemma:conv_prop} establishes that if the true belief state $\pi_\tau$ is available,  then the optimal value function can be obtained by Algorithm \ref{Alg1}.  {Accordingly, the optimal prescription  can be approximated by  $\hat{\gamma}^*_t={\argmax_{\delta\in\Gamma}V(h_t\delta)}$, which yields  the approximate optimal control action $\hat{\gamma}^{i,*}_t(m^i_t)$.}  

\subsection{Connections to Existing Solvers}\label{sec:connection}

The common information approach advocated in the control community has connections to some Dec-POMDP solvers from the computer science  community. For instance, one popular heuristic algorithm for solving Dec-POMDPs is the tree search algorithm MAA$^*$ \cite{oliehoek2014dec}. 
In the context of MAA$^*$, each node in the tree denotes {a} history of joint policies, which can be interpreted as  {a} state; and each edge represents a joint decision rule, which corresponds to  {an} action. The authors formulate the problem as {a NOMDP} \cite{oliehoek2014dec}, defining a sufficient statistic as the distribution over joint observation histories and states. Drawing a connection to our approach, this reduction can be obtained via a special case of the common information approach, where the common information is empty and the local memory is the local observation history. In this sense, our algorithm can be viewed as a generalization of the Dec-POMDP solver presented in \cite{oliehoek2014dec}.



The common information approach is also related to the reduction of Dec-POMDPs to \emph{occupancy state MDPs} \cite{dibangoye2016optimally}, where the occupancy state is defined as the {joint distribution} over states and the joint history of actions and observations. 
As noticed in \cite{dibangoye2016optimally}, the reduction to a  {NOMDP} is a special case of the occupancy state reduction when deterministic policies are used. The latter {approach} is then included as a special case of the common information approach when the local memory is the history of local actions and observations and the common information is null. While the occupancy state reduction is amenable to sampling-based approaches, \cite{dibangoye2016optimally} makes use of the piece-wise linearity and convexity of the optimal value function to solve the problem.

%% file: Example.tex
\section{Application: Collaborative Intrusion Response in Cyber Networks}\label{sec:Example}

We consider the problem of \emph{collaborative intrusion response} describing how a collection of defenders can collaboratively achieve system-wide security under the constraint that each agent can only prescribe localized defense actions based on localized security alert information. Our setting goes one step beyond \emph{collaborative intrusion detection systems} \cite{vasilomanolakis2015taxonomy} by addressing the question of not only attack detection, but attack response.

Following existing work, we model the cyber network by a type of attack graph termed a condition dependency graph \cite{ammann2002scalable,miehling2018pomdp}. The dependency graph, denoted by $
\cG = \{\cS,\cE\}$, quantifies the relationship between security conditions (attacker capabilities), represented by nodes $\cS$, and exploits, represented by hyperedges $\cE$. Specifically, each node in $\cS$ is assumed to either be enabled (attacker possesses the capability) or disabled (attacker does not possess the capability) whereas each edge $e_j\in\cE$ is an ordered pair of sets, $e_j = (\Npre_j,\Npost_j)$, relating the conditions necessary for the exploit to be attempted, termed \emph{preconditions} $\Npre_j\subseteq\cS$, to conditions that become enabled if the exploit succeeds, termed \emph{postconditions} $\Npost_j\subseteq\cS$. 
The system state, termed a \emph{security state}, is defined as the set of currently enabled nodes. We consider a simple probabilisitic threat model. For a given security state, the attacker is attempts exploits with enabled preconditions $e_j$ with a fixed probability $\alpha_j$, where each attempted exploit $e_j$ succeeds with a fixed probability $\beta_j$. 
Defense actions induce system modifications that have the effect of blocking certain exploits from succeeding (setting $\beta_j=0$ for each blocked exploit). 
Each agent $i$ is associated an intrusion detection system which generates security alerts from the set $\cA^i$. Attempt of an exploit $e_j\in\cE$ generates alert $k$ with probability of detection $\delta_{jk}$. Additionally, each alert $a_k$ is also subject to false alarms as dictated by the (per time-step) probability $\zeta_k$.
%
Lastly, the attacker's goals (enabling specific nodes termed \emph{goal conditions}) are encoded by a cost function which takes into account the tradeoff between security (keeping the attacker away from its goal conditions) and availability (preserving network usability by limiting system modifications).





The defining feature of the security problem is that no agent possesses system-wide knowledge. Agents have asymmetric information over the factors that influence their decision-making; the evolution of the security of the system, as well as the cost of an agent's defense decision, depend on all agents' defense actions which are not known by other agents. Furthermore, inference of the security status of the system depends on security alerts from all IDSs; however, each agent only receives local alerts from their own IDS. 

We assume that agents can share such information (defense decisions and security alerts) via a centralized database; however, due to practical limitations, the updating of this database cannot be done instantaneously. There is an inherent delay in updating the information. 
The delayed sharing information structure \cite{witsenhausen1971separation,nayyar2013decentralized}, a special case of the general model discussed in this paper, formalizes this interaction. As described in Section \ref{ssec:com}, the general form of each agent's belief is a joint distribution over the underlying state and the local information of others (its unknown information). In this example, under the delayed sharing information structure, each agent's belief is the joint distribution over the security state and the information (joint defense actions and security alerts) that is not yet available in the central database. 

\subsection{A Small Example}

\begin{figure}[t]
\begin{center}
\includegraphics[width=0.925\columnwidth]{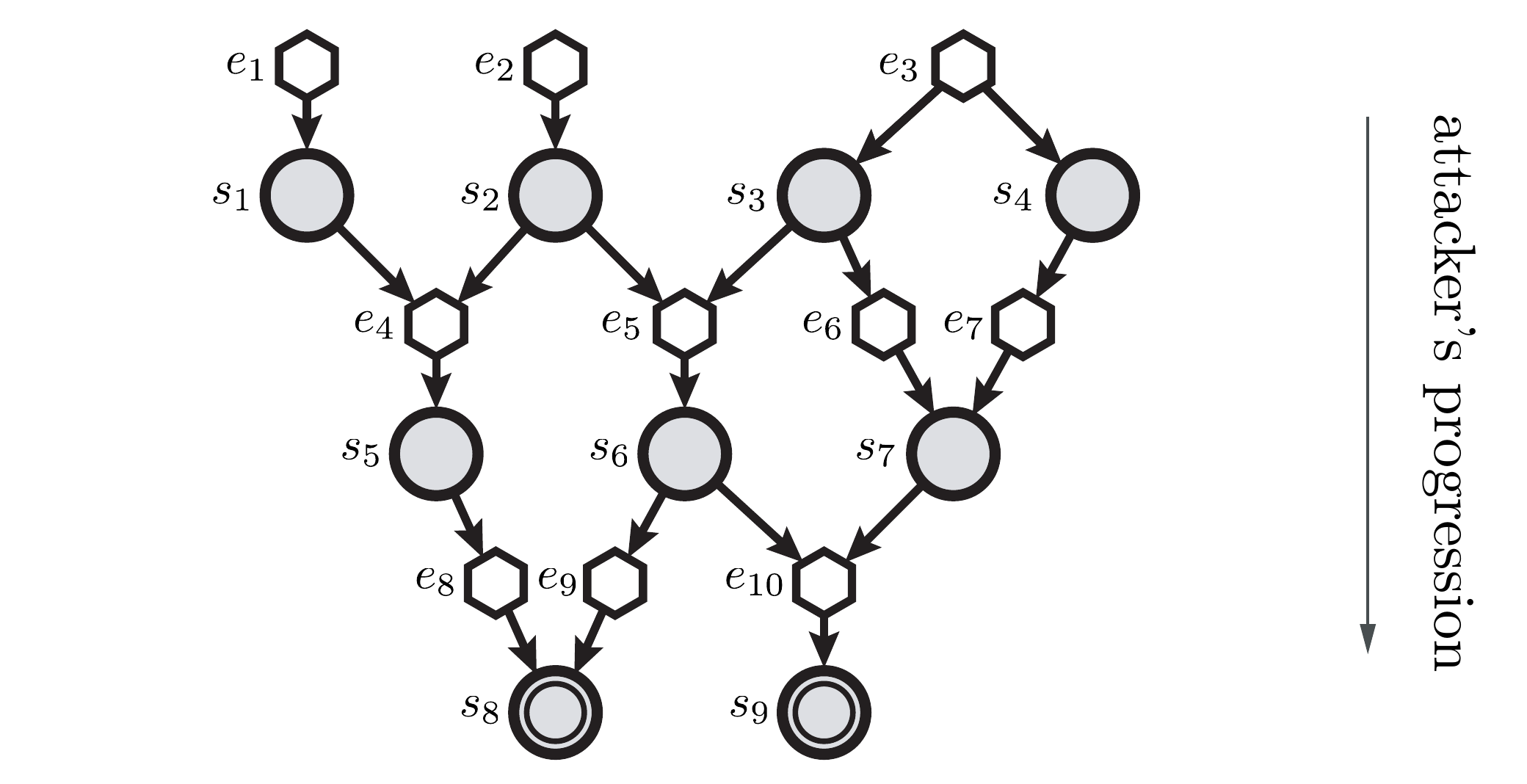}
\caption{\small \emph{The dependency graph of the example}. The graph consists of ten exploits (\emph{e.g.} $e_1=(\varnothing,\{s_1\})$, $e_4=(\{s_1,s_2\},\{s_5\})$, \emph{etc.}) and nine security conditions, with goal conditions $s_{8}$ and $s_{9}$ represented by double-encircled nodes.}
\label{fig:cdg}
\end{center}
\vspace{-3mm}
\end{figure}

We study a small instance of the collaborative intrusion response model consisting of $n=2$ agents. The cyber network is represented by the condition dependency graph of Fig. \ref{fig:cdg}. Each agent can control the status of a set of exploits; agent $1$ controls $e_1,e_2,e_3,e_4,e_6$ and agent $2$ controls $e_5,e_7,e_8,e_9,e_{10}$. Each agent's action space is $\uset^i=\{0,1\}$, where $1$ (resp. $0$) corresponds to all exploits under its control being blocked (resp. not blocked). The threat model is described by uniform probabilities of attack and success, $\alpha_j=\beta_j=0.5$ for all $e_j\in\cE$. Each agent's IDS generates a single alert, $\cA^1=\{a_1\}$, $\cA^2=\{a_2\}$, with $\cA(e_j)=a_1$, $j=1,2,\ldots,7$, and $\cA(e_j)=a_2$, $j=4,5,\ldots,10$, with probabilities of detection $\delta_{j1}=0.8$ for $j=1,2,3$; $\delta_{j1}=0.1$, $\delta_{j2}=0.3$ for $j=4,5,6,7$; $\delta_{j3}=0.8$ for $j=8,9,10$, zero otherwise. Probabilities of false alarm are $\zeta_1 = \zeta_2 = 0.3$. Each agent's observation space is defined as $\yset^i=\{0,1\}$ designating the presence of an alert from its IDS. The instantaneous cost function $c:\cS\times\cU\to\RR$ is $c(s,u) = c(s)+c(u)$, where $c(s)=5$ if $s_8,s_9\in s$, zero otherwise, and $c(u) = 0$ for $u=(0,0)$; $c(u) = 1$ for $u=(0,1)$, $u=(1,0)$; and $c(u) = 4$ for $u=(1,1)$.


We assume that defense actions and security alerts are subject to a 1-step delay before becoming available in the central database (common history). Thus, the local memory at time $t$ of each agent $i$ is given by $m_t^i = \{u_{t-1}^i,y_t^i\}\in\cM_t^i = \uset^i\times\yset^i = \{0,1\}^2$. The information in the database at time $t$ is $c_t = \{u_{1:t-2},y_{1:t-1}\}$ whereas innovations are given by $z^i_{t+1} = \protcom^i(m_t^i,u_t^i,y_{t+1}^i) = m_t^i = \{u^i_{t-1},y^i_t\}$, with joint innovation $z\in\cZ=\uset\times\yset = \{0,1\}^2\times\{0,1\}^2$, resulting in $c_{t+1} = \{c_t,z_{t+1}\} = \{u_{1:t-1},y_{1:t}\}$. Local memories are updated as $m_{t+1}^i = \protloc^i(m_t^i,u_t^i,y_{t+1}^i,z_{t+1}^i) = \{u_t^i,y_{t+1}^i\}$. The number of prescriptions is $|\Presc_t| = \prod_{i=1}^2|\cU^i|^{|\locset_t^i|} = \prod_{i=1}^2|\cU^i|^{|\cU^i\times\cY^i|} = 2^4\cdot2^4 = 256$ for all $t$.

\subsection{Numerical Results}

\begin{figure}[t]
\begin{center}
\includegraphics[width=0.9\columnwidth]{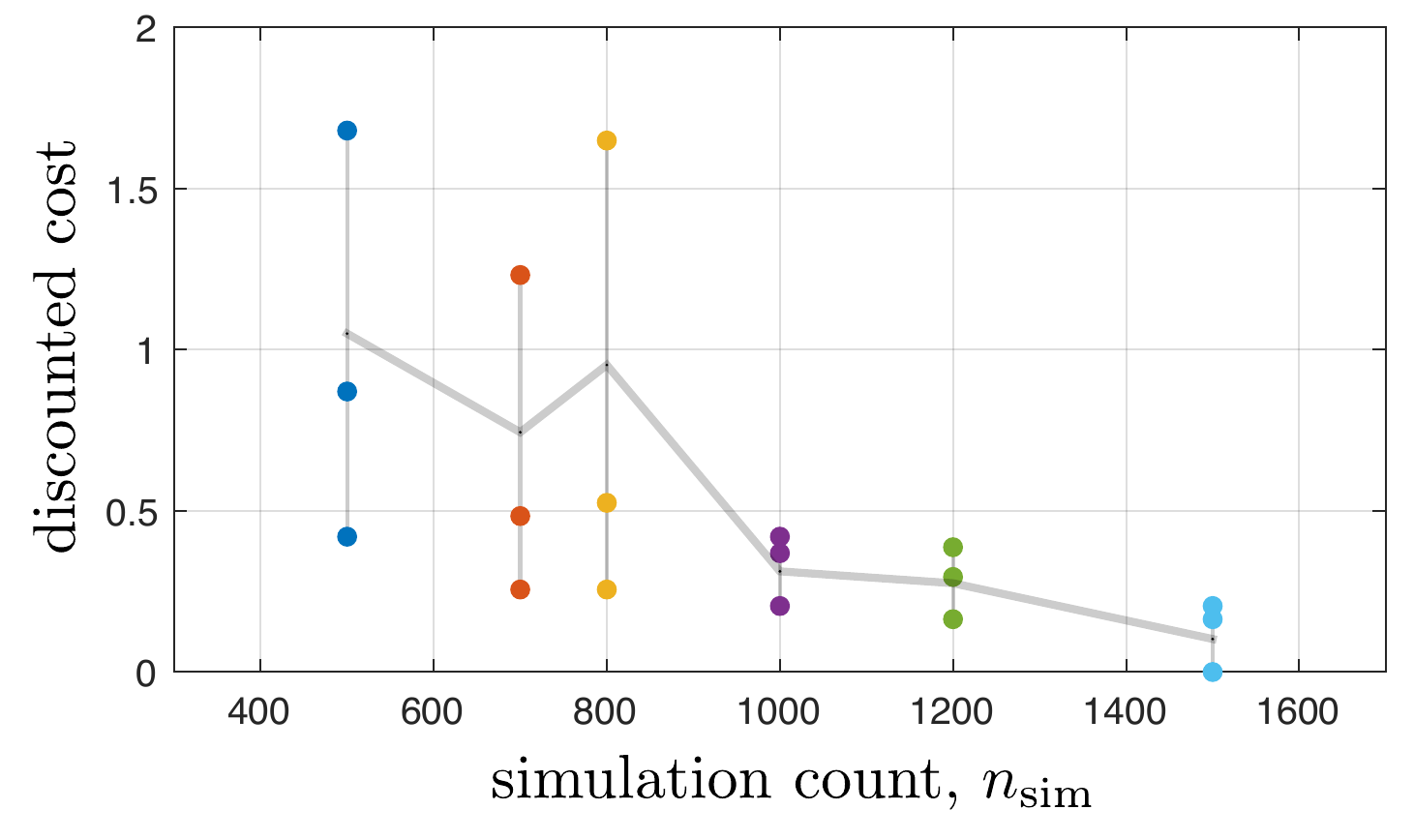}
\caption{\small Discounted cost for a given time-step ($t=5$) under various simulation counts. Each point represents the cost at $t$ for a distinct sample path for the given value of $n_{\text{sim}}$. The gray line represents the mean discounted cost.}
\label{fig:costplot}
\end{center}
\vspace{-3mm}
\end{figure}

We investigate the quality of the resulting policy, as computed by Algorithm 1, as a function of the number of simulations used to compute each action, denoted by $n_{\text{sim}}$. We assume a discount factor $\beta = 0.8$, discount horizon threshold $\varepsilon = 0.1$, exploration constant $\rho = 10$, particle count $K=400$, and a uniform random rollout policy. The performance of the computed policy is illustrated in Fig. \ref{fig:costplot}.
The figure empirically verifies the convergence result in Section \ref{sec:Analys}, \emph{i.e.}, the discounted cost decreases as $n_{\text{sim}}$ increases. 


%% file: DiscussionAndConclusion.tex
\section{Discussion and Concluding Remarks}\label{sec:disc}

In this paper, we have proposed an online {tree-search based} algorithm for obtaining team-optimal policies in decentralized stochastic control problems {when} agents share some of their history with others. 
{Our algorithm enables each agent to obtain approximately optimal control policies without explicit communication or model knowledge. 
We have also shown that two recent algorithms that solve Dec-POMDPs \cite{oliehoek2014dec,dibangoye2016optimally} can be viewed as special cases of our algorithm. Lastly, we have demonstrated the performance of our algorithm in a novel computer security setting. 
}

We are not the first to consider the applicability of tree-search methods in multi-agent decision environments. {In fact, tree-search methods do not scale well for decentralized problems due to the large joint action/observation spaces in the multi-agent setting \cite{amato2015scalable}. That said, by leveraging common information, our algorithm serves as an initial step towards developing tractable sampling-based planning/learning algorithms for decentralized stochastic control problems, especially without full knowledge of the system model.   
}

}




